\documentclass[twocolumn]{article}

\usepackage{arxiv}
\usepackage{placeins}
\usepackage[utf8]{inputenc}
\usepackage[T1]{fontenc}
\usepackage{hyperref}
\usepackage{url}
\usepackage{booktabs}
\usepackage{amsfonts}
\usepackage{nicefrac}
\usepackage{microtype}
\usepackage{cleveref}
\usepackage{graphicx}
\usepackage[numbers,sort&compress]{natbib}
\usepackage{doi}
\usepackage{graphicx}
\usepackage{adjustbox}
\usepackage{booktabs}
\usepackage{multirow}
\usepackage[normalem]{ulem}
\usepackage{xcolor}
\usepackage{wrapfig}
\usepackage{lmodern}
\usepackage[T1]{fontenc}
\usepackage{threeparttable}
\usepackage{makecell}   
\usepackage[table]{xcolor}
\usepackage{caption}
\definecolor{bestgreen}{RGB}{198,239,206}
\definecolor{secondorange}{RGB}{255,235,156}
\definecolor{lightgreen}{RGB}{232,247,235}
\definecolor{lightorange}{RGB}{255,246,210}
\usepackage{caption}
\usepackage{fvextra}
\DefineVerbatimEnvironment{codeblock}{Verbatim}{
  breaklines,
  breakanywhere,
  fontsize=\small,
  frame=lines,
  framesep=2mm,
  baselinestretch=1.05
}
\usepackage{siunitx}    
\usepackage{enumitem}
\newcommand{\methodname}{\textsc{Void}}
\title{
\raisebox{-0.55em}{\includegraphics[height=2.3em]{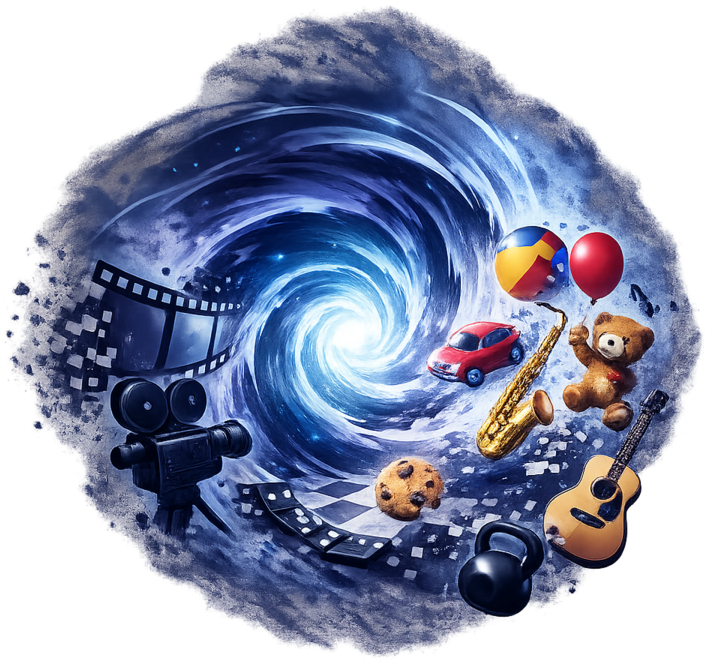}}
VOID: Video Object and Interaction Deletion
}
\author{
\normalfont
Saman Motamed$^{1,2}$, William Harvey$^{1}$, Benjamin Klein$^{1}$,\\
Luc Van Gool$^{2}$, Zhuoning Yuan$^{1}$, Ta-Ying Cheng$^{1}$\\
[0.3em]
$^{1}$Netflix \quad
$^{2}$INSAIT, Sofia University ``St.\ Kliment Ohridski''\\
[0.8em]
\textcolor{magenta}{\url{https://void-model.github.io}}
}
\newcommand{\mo}{\mathbf{M}_o}
\newcommand{\mq}{\mathbf{M}_q}
\newcommand{\ma}{\mathbf{M}_a}
\newcommand{\vid}{\mathbf{V}}
\newcommand{\vidhat}{\hat{\mathbf{V}}}

\begin{document}

\makeatletter
\twocolumn[{%
\begin{@twocolumnfalse}
\maketitle

\centering
\includegraphics[width=\textwidth]{teaser-fin.jpg}
\captionof{figure}{Removing an object and its interactions can require rewriting the entire scene.
    On the left, when the middle three blocks are removed, VOID correctly models the domino effect halting so that the yellow block never falls.
    On the right, when the hands are removed, VOID correctly models the spinning tops continuing without interruption.}
\label{fig:teaser}

\begin{abstract}
Existing video object removal methods excel at inpainting content ``behind'' the object and correcting appearance-level artifacts such as shadows and reflections.
However, when the removed object has more significant interactions, such as collisions with other objects, current models fail to correct them and produce implausible results.
We present \methodname{}, a video object removal framework designed to perform physically-plausible inpainting in these complex scenarios. 
To train the model, we generate a new paired dataset of counterfactual object removals using Kubric and HUMOTO, where removing an object requires altering downstream physical interactions.
%
%
During inference, a vision-language model identifies regions of the scene affected by the removed object. These regions are then used to guide a video diffusion model that generates physically consistent counterfactual outcomes.
%
%
Experiments on both synthetic and real data show that our approach better preserves consistent scene dynamics after object removal compared to prior video object removal methods. 
%
%
We hope this framework sheds light on how to make video editing models better simulators of the world through high-level causal reasoning.
\end{abstract}

\vspace{0.5em}
\end{@twocolumnfalse}
}]
\section{Introduction}
\label{sec:intro}
Videos capture the complex causal dynamics of our physical world. The presence of an object exerts a diverse range of effects on its environment, ranging from photometric phenomena like shadows and reflections to kinetic events like collisions. This spatiotemporal entanglement makes the seemingly simple task of ``video object removal'' non-trivial for video generation and editing models. It requires the model to first imagine ``What would happen if this object was removed?''  and then synthesize a realistic video of its answer. Consider the line of collapsing domino tiles shown on the left of \cref{fig:teaser}. If we use a video inpainting model to remove the middle tiles, the later tiles keep falling, which is a physically impossible scenario. A realistic solution requires the model to reason that without the middle tiles, tiles appearing later in the chain must remain standing. Doing so requires an editing model to simulate the world through high-level causal reasoning and not rely solely on low-level visual features. Mastering this capability will benefit film visual effects and make advanced video editing accessible to non-experts.

Video decomposition methods \cite{lu2021omnimatte, lee2024generative, samuel2025omnimattezero,shrivastava2024video} aim to decompose videos into layers, such that each object has an associated layer containing its effects disentangled from those of other objects in the scene. These methods excel at extracting effects where an object creates shadows, reflections, or other distinct visual entities. However, they are not capable of disentangling interactions where one object affects another, such as by breaking or moving it. Alternatively, a text-guided video editing model~\cite{runway_gen4_2025, jiang2025vace, kulikov2026versatile} could remove an object and its effects following a user prompt. However, current diffusion models often lack the physical intuition to determine what should happen with the object removed, and it is often impossible to specify all effects precisely with a text prompt.

To this end, we propose an extension of video object removal to more dynamic scenarios. These require not only removing a specified object, but also modeling how its removal affects other objects in the scene. We then present \methodname{}, a framework to elicit this high-level causal reasoning from a video diffusion model.

\methodname{} is built on advances along three axes: data construction, training strategy, and inference optimization. To create data pairs that capture dynamic changes when an object is removed, we repurpose the Kubric simulation and rendering engine~\cite{greff2022kubric} and the HUMOTO human motion capture dataset~\cite{humoto}. For training, we propose two improvements over prior work~\cite{lee2024generative}: (i) ``quadmask'' conditioning that explicitly identifies regions of each frame that may change after the object is removed, and (ii) a video appearance refiner applied in a second pass to remove artifacts like unwanted object morphing. During inference, we generate quadmasks with vision-language models (VLMs), leveraging their world knowledge to expand a simple object mask into richer pixel-space guidance.

We gather a new benchmark of videos with diverse and complex interactions, comprising synthetic and real-world data. Extensive studies involving perceptual metrics, user studies, and VLM-as-a-judge demonstrate compelling results from \methodname{} against both video inpainting methods and general video editing models. We further see surprising generalizations to unseen effects: VOID models a balloon floating up when the person holding it is removed, despite there being no floating objects in its training data; VOID also prevents food inside a blender from moving when the person turning it on disappears, despite there being no blenders or electrical devices in its finetuning data. 
These extrapolations demonstrate that \methodname{} does not just recall simple visual cues from its training data, but applies high-level reasoning and world knowledge from the VLM and underlying video diffusion model to video editing.  As such, it is likely to benefit as more capable generative models become available.

In summary, we have three contributions. First, we investigate the current pitfalls of physics-aware object removal when the removed object has complex interactions with the rest of the scene. Second, we introduce \methodname{}, a framework that tackles these problems from three perspectives: data curation, training strategy, and inference time VLM-guided scene analysis. Finally, we present extensive evaluations on previous and new benchmarks that show the superiority of our model in disentangling and removing a wide range of object effects. We believe that this work illuminates an interesting and underexplored direction for video generation and world modeling research.


\section{Related Work}
\label{sec:related}

\paragraph{\textbf{Video Generation and Editing.}}
The breakthrough of diffusion and flow-matching methods has led to large advancements in video generation models. Notable examples include closed-source models like Veo 3~\cite{deepmind2025veo3} and Runway~\cite{runway_gen4_2025} and open-source models such as WAN~\cite{wan2025}, VACE~\cite{jiang2025vace}, CogVideo~\cite{cogvideo}, and LTX-2~\cite{ltx2_2026}. Much of the stunning performances of these models came from the large video-text datasets driving forward the performance, allowing them to extend to various editing controls including text and sketches. However, since they are learning from unstructured data, they often create pixel-perfect yet physically implausible scenes, especially for tasks that require extensive reasoning (e.g., removing an object).

Several models aim to improve reasoning via VLMs~\cite{yunhe24langdrivectrl,veggie2025,videorepair2024}. However, these models either work in a specific domain (e.g., LangDriveCtrl~\cite{yunhe24langdrivectrl} works purely for driving scenes) or only solve simple reasoning tasks like grounding (Video-Repair~\cite{videorepair2024}) and segmentation (Veggie~\cite{veggie2025}). \methodname{} takes a leap towards applying VLM reasoning for complex video editing tasks, where we need to synthesize counterfactual scenarios in which an object is removed.

\paragraph{\textbf{Video Decomposition and Effect Removal.}}
The problem of decomposing a video into RGBA layers was significantly advanced by Omnimatte \cite{lu2021omnimatte}, which introduced a self-supervised framework to associate subjects with their ``effects'' (e.g., shadows and reflections). While OmnimatteRF \cite{kim2023omnimatterf} extended this by modelling the static background with 3D radiance fields, these methods remain fundamentally reconstructive, focusing on uncovering existing background pixels rather than synthesizing new content. More recently, Generative Omnimatte~\cite{lee2024generative} integrated priors from a video inpainting model, using a trimask setup to decompose images into object-specific layers and effects. OmnimatteZero~\cite{samuel2025omnimattezero} introduced training-free extraction of effects and objects via attention maps. 

There are also a plethora of recently-released video inpainting methods. Propainter \cite{zhou2023propainter} proposed a dual-domain propagation from image and feature side for better inpainting. DiffuEraser~\cite{li2025diffueraser} integrated flow-based pixel propagation with transformer-based generation to better restore textures and objects. AVID~\cite{zhang2024avid} and FDM~\cite{green2024semantically} proposed sampling pipelines to extend video inpainting lengths. Minimax-Remover~\cite{zi2025minimax} is an object removal approach with a more efficient model architecture followed by distilling a remover on human annotations. ROSE~\cite{rose2025} proposed an effect removal inpainting framework focusing on photometric effects such as shadows, reflections, light, and translucency, while Object-Wiper~\cite{objectwiper2026} presented a training-free method targeting these effects. While these methods all handle some associated photometric effects of the removed object, they cannot model complex physical interactions.


\section{Approach}
\label{sec:approach}

\begin{figure*}[t]
    \centering
    \includegraphics[width=\linewidth]{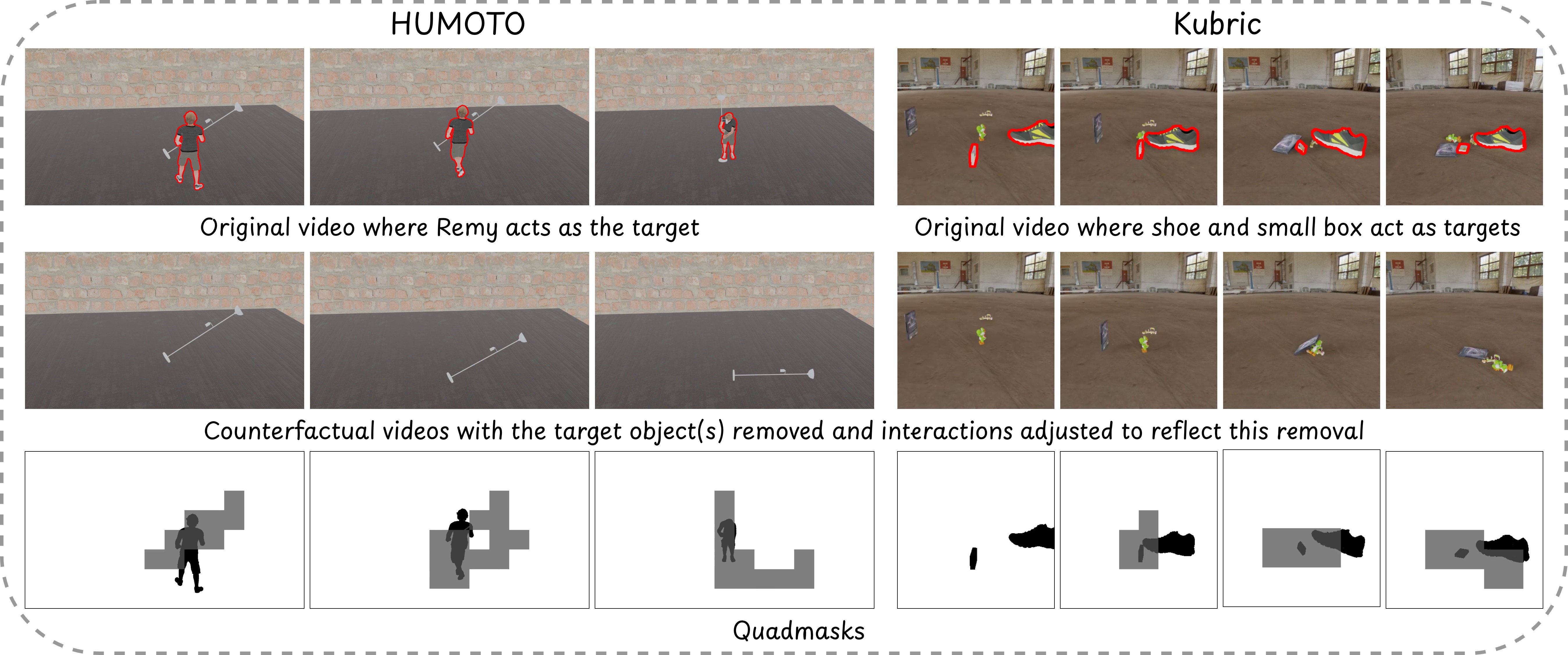}
    \caption{\textbf{Counterfactual supervision examples.}
    Top: videos $\vid$ where $O$ is outlined in red.
    Bottom: re-simulated counterfactuals $\vidhat$ generated without $O$.
    In Kubric scenes, downstream motion changes when the initiating object is removed.
    In HUMOTO scenes, supported objects transition naturally under gravity.}
    \label{fig:dataset_samples}
\end{figure*}

\begin{figure*}[t]
    \centering
    \includegraphics[width=\linewidth]{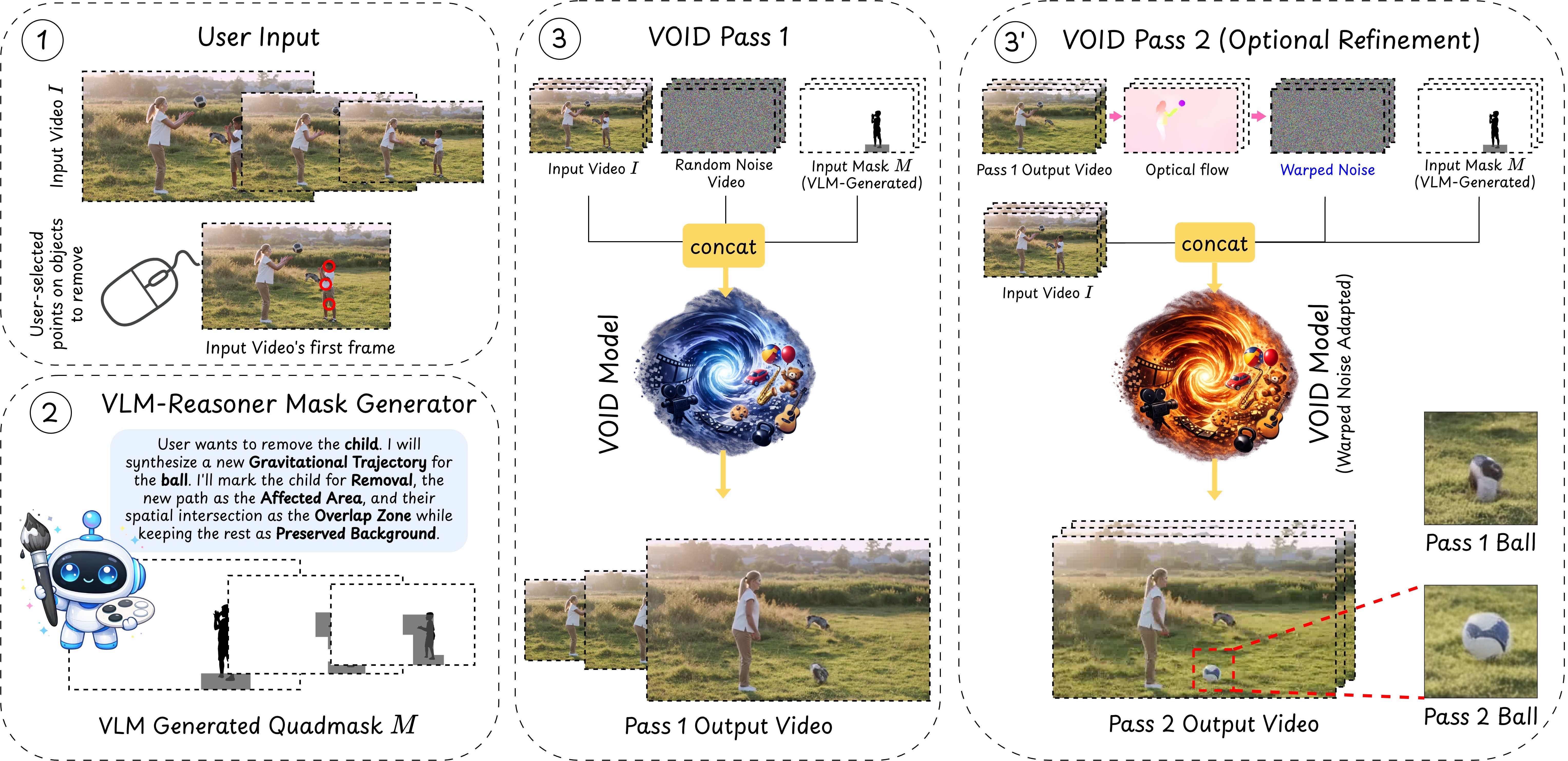}
    \caption{\textbf{\methodname{}: Interaction-Aware Counterfactual Video Generation.} A user provides an input video and clicks on an object to mask it for removal. A VLM-based pipeline expands the mask to identify other areas that will be affected. VOID's first pass then predicts a counterfactual trajectory. The optional second pass stabilizes object deformation using flow-warped noise derived from the initially predicted motion.}
    \label{fig:void-method}
\end{figure*}

We start from an input video $\vid = \{I_t\}_{t=1}^T$ and a mask sequence $\mo = \{m_t\}_{t=1}^T$ identifying one or more target objects to remove, $O$. Our objective is to learn a model $f$ that generates a counterfactual video $\vidhat$, in which $O$ and all induced interactions are removed:
\begin{equation}
\vidhat = f(\vid, \mo).
\end{equation}
In general, the interactions can be complex. Removing a support can cause something to fall. Removing an obstacle can prevent a collision. To work well in these settings, $f$ cannot rely on spatial hole filling but must conceptualize how the scene would have evolved in the target object's absence. It should then \textit{(i)} eliminate the target object, \textit{(ii)} regenerate regions affected through potentially complex relationships, and \textit{(iii)} preserve unaffected regions.


\subsection{Counterfactual Dataset Supervision}
\label{sec:data}

Training this model requires a dataset of counterfactual video pairs $(\vid, \vidhat)$ with and without object $O$, respectively. Video pairs used by existing video inpainting and omnimatte datasets~\cite{rose2025, lee2024generative, xu2018youtube, davis} focus mostly on photometric effects such as shadows and lack supervision for the removal of objects that physically affect other objects in the scene. We generate new counterfactual pairs to address this gap with physics-based simulations from Kubric~\cite{greff2022kubric} and human motion capture data from HUMOTO~\cite{humoto}.
%
For both datasets, we randomize camera trajectories and focal zoom during rendering to help with the disentanglement of object effects and camera trajectories.
We show in \cref{fig:more-results} that training on these two synthetic datasets enables extensive generalization to real-world domains.

\paragraph{\textbf{Rigid-body dynamics (Kubric).}}
The diverse set of objects offered by Kubric is ideal for simulating collisions, falling, and structural dependencies.  We create videos $\vid$ by sampling initial conditions of multiple objects with varying initial positions and velocities, and simulating the interactions over time. We then define one or more of the objects to be $O$. The counterfactual video $\vidhat$ is created by removing $O$ while keeping initial conditions for all other objects the same, and re-simulating the scene. This yields a new, physically consistent alternative set of interactions. We generate $\sim$1900 videos pairs in this manner.

\paragraph{\textbf{Articulated interactions (HUMOTO).}}
We use HUMOTO, a 4D motion-capture dataset of human-object interactions, to collect articulated interaction data. 
In these sequences, $O$ corresponds to the human performing diverse activities. $\vid$ and $\vidhat$ are created by passes over the simulation and rendering engine with and without the human.
These videos teach the model how to perform object removal in videos containing dynamic manipulations. We randomize the textures of the objects in the scene, the background wall and the human, and generate $\sim$4500 video pairs. 


\subsection{Interaction-Aware Quadmask Conditioning}
\label{sec:quadmask}
To provide further guidance over the binary object mask $\mo$, Lee et al.~\cite{lee2024generative} proposed a trimask which distinguishes between three image regions: the object to be removed (black), the area affected by its removal (light gray), and areas which should stay the same (white). Their setup, however, creates two ambiguities. 

The first is that they highlight almost the entire region of each image frame in light gray and mark only specific objects as white. Their model therefore learns that it typically needs to modify only a small portion of the light gray mask to remove the effects.
%
%
We provide stronger guidance by focusing the light gray region closely on where effects take place. While generating data, we use the rendering engines to determine these regions. We then gridify the regions to better match our inference time procedure described in \cref{sec:vlm_inference}.

The second ambiguity occurs when there is overlap between the object to remove and the area with dynamic effects. In the example in \cref{fig:void-method}, we want to remove a child and once they are removed, the ball should fall to the ground instead of them catching it. Consider what value the trimask should have around the boy's upper body while he catches the ball. Should it be black because the boy is being removed? Or should it be light gray because the ``effect'' of the removal is that the ball should now continue its trajectory and pass through this area? To resolve these ambiguities, we extend the trimask to a quadmask $\mq$ with a fourth color (dark grey) that describes overlap between (i) the object to be removed and (ii) other parts of the scene that are affected. See \cref{fig:dataset_samples} for examples.

\subsection{Backbone Initialization and Counterfactual Generation}
\label{sec:backbone}

We propose \methodname{}, a model built upon the CogVideoX diffusion transformer backbone~\cite{cogvideo} and initialized from the weights released with Generative Omnimatte~\cite{lee2024generative}. This initialization provides a strong prior for layered object–effect disentanglement under trimask guidance. We finetune it with quadmask conditioning on the counterfactual video pairs described previously. This teaches the model mask semantics and re-enables the underlying video model's native capacity for physically plausible trajectory synthesis, transforming it from a layered removal model to a dynamic counterfactual rewriting model.

\subsection{Pass~1: Counterfactual Trajectory Synthesis}
\label{sec:pass-1}
In its first pass, \methodname{} generates an initial counterfactual prediction:
\begin{equation}
\vidhat_\text{p1} = \text{VOID}(\mathbf{z}, \vid, \mq),
\end{equation}
where $\mathbf{z} \sim \mathcal{N}(\mathbf{0}, \mathbf{I})$ denotes the Gaussian diffusion noise,
$\vid$ is the input video sequence, and $\vidhat_\text{p1}$ is \methodname{}'s first-pass prediction of how the scene would evolve in the absence of the target object. This pass typically captures broadly correct hypotheses around motion, such as previously supported objects entering free-fall and previously-obstructed objects continuing their motion. However, we find that the objects undergoing newly synthesized motion can exhibit structural deformation.

\subsection{Pass~2: Flow-Warped Noise Stabilization}
\label{sec:deformation}

\begin{figure}
    \centering
    \includegraphics[width=\linewidth]{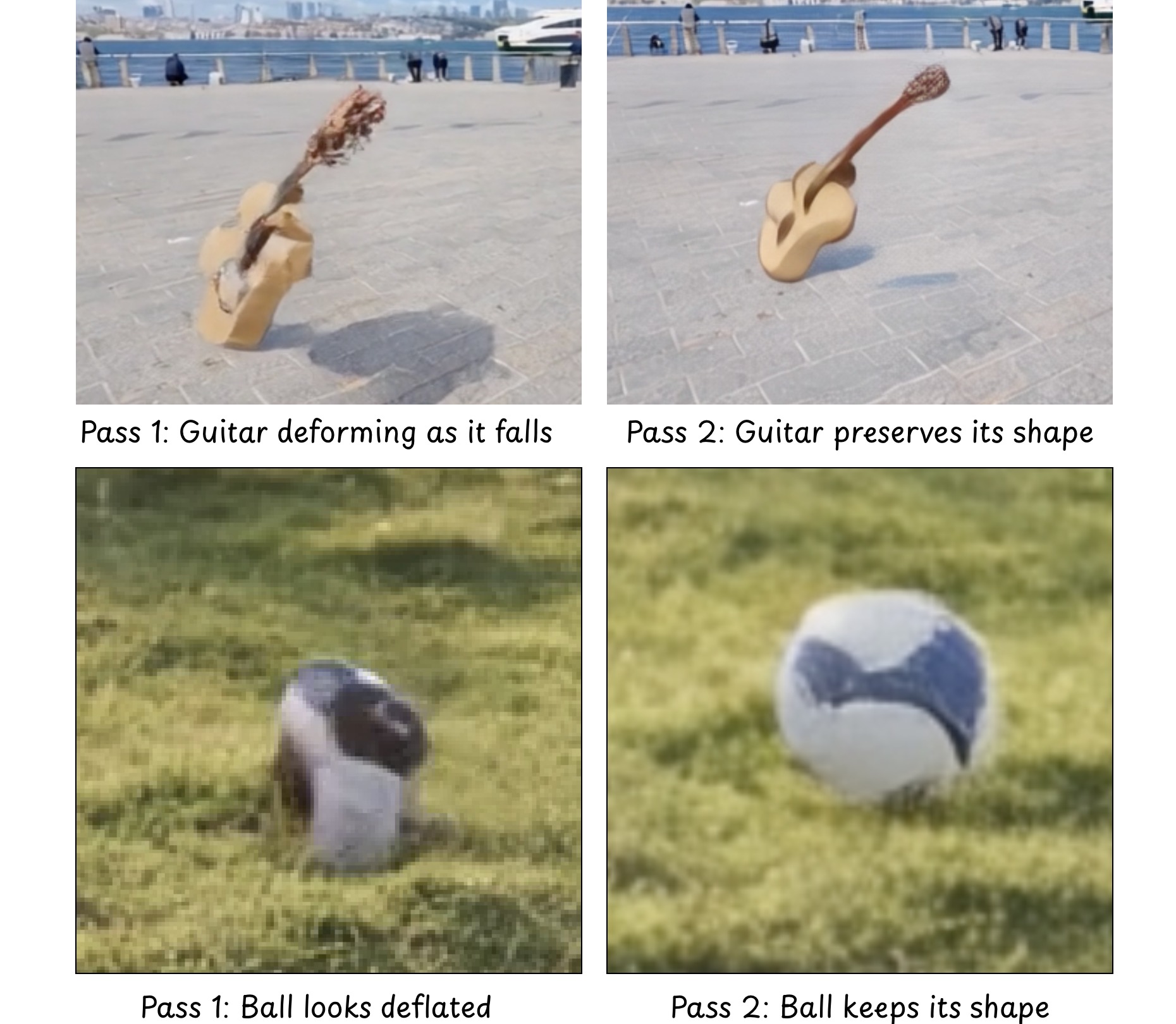}
    \caption{
    Frames from generated videos featuring a guitar entering free-fall and a thrown ball following a new trajectory.
    Pass~1 (left) produces correct counterfactual trajectories but exhibits structural deformation.
    Pass~2 (right) better preserves object rigidity by using motion-aligned warped noise.
}
    \label{fig:pass1_vs_pass2}
\end{figure}
\paragraph{\textbf{Why deformation occurs.}}
Video diffusion models, especially relatively lightweight models like the 5 billion parameter CogVideoX model we build on, struggle to maintain temporal coherence when generating complex, motion-heavy videos~\cite{chefer2025videojam,motamed2025generative}. Prior work mitigates this issue through large-scale tracking supervision or explicit motion conditioning~\cite{chefer2025videojam,Motamed_2024_CVPR,jeong2025track4gen,gu2025diffusion,liu2024intragen, boduljak2026what}. In the simple object removal case where only photometric effects need to be corrected, the input video provides similarly strong constraints on the generated motion. For example, if we need to remove a shadow from a surface, the motion and geometry of the surface in the output video should be the same as in the input. In our more complex settings, the diffusion model often needs to generate new motion. We find that this leads to bending, stretching, or structural drift of the objects undergoing changed motion, similar to the artifacts seen from running the CogVideoX image-to-video model without motion guidance. We now describe how to resolve these artifacts in the object removal setting without changing the base model. 




\paragraph{\textbf{Second pass to fix deformation}}
Go-with-the-Flow~\cite{burgert2025gowiththeflow} observed that using temporally correlated noise based on predicted motion trajectories can encourage the diffusion model to denoise consistently along those trajectories. We follow Go-with-the-Flow~\cite{burgert2025gowiththeflow} to derive warped noise from the optical flow field of our first-pass output $\vidhat_\text{p1}$, and then use it as input to a second pass as:
\begin{equation}
\vidhat =
\text{VOID}_\text{warp}(\mathbf{z}_\text{warp}, \vid, \mq),
\end{equation}
where $\text{VOID}_\text{warp}$ is a warped noise variant of \methodname{}. It is trained with the same data and quadmask conditioning but with flow-aligned noise derived from each training target $\vidhat$.

This second pass is not always required so we trigger it only when object removal is predicted to cause substantial dynamic reconfiguration. The same VLM used to create a quadmask additionally classifies whether removal induces significant object motion (e.g., free-fall or trajectory change). We trigger the second pass only when such dynamics are detected. Figure~\ref{fig:pass1_vs_pass2} shows the effect of pass 2 on two objects.

\subsection{VLM-Guided Quadmask Generation at Inference Time}
\label{sec:vlm_inference}

At inference time, we start from an input video $\vid$ and user-provided binary object mask $\mo$. To run \methodname{}, we need to first infer the affected region and use it to create the quadmask $\mq$.
Doing so requires reasoning about counterfactual dependencies and object dynamics, so we use a VLM~\cite{chow2025physbench, motamed2025travl, li2025survey}. We start by inputting $\vid$ and $\mo$ to the VLM and using it to produce a list of descriptions of objects that are affected by the removed object. We use Segment Anything 3~\cite{carion2025sam3segmentconcepts} to get a mask $\ma^\text{orig}$ covering all objects in the list. Since the affected objects may be in different places in the counterfactual scenario, we also need to predict their counterfactual positions to fully capture the changes between $\vid$ and $\vidhat$. We therefore feed $\ma^\text{orig}$ into the VLM and use it to predict the mask sequence describing the positions of these objects in the counterfactual scenario. This is done by overlaying a coarse spatial grid on the input video and asking it to list which cells in each frame may contain effects. This gives us a block-structured mask $\ma^\text{count}$ describing where the affected objects in $\ma^\text{orig}$ go in the counterfactual scenario. We combine the masks to get the final affected area mask $\ma := \ma^\text{orig} \lor \ma^\text{count}$. We finally compute the quadmask $\mq$ by setting it to black for pixels only in $\mo$; dark grey for pixels where $\mo$ and $\ma$ overlap; light grey for pixels only in $\ma$; and white everywhere else.

\section{Results}
\label{sec:results}

We test on two datasets. The first comprises 75 real-world videos involving object manipulation, support removal, collisions, articulated interactions, and shadow/reflection removal. The second is synthetic and consists of 30 Kubric and HUMOTO test videos combined with existing synthetic object removal datasets.

\begin{figure*}[ht!]
    \centering
    \includegraphics[width=\linewidth]{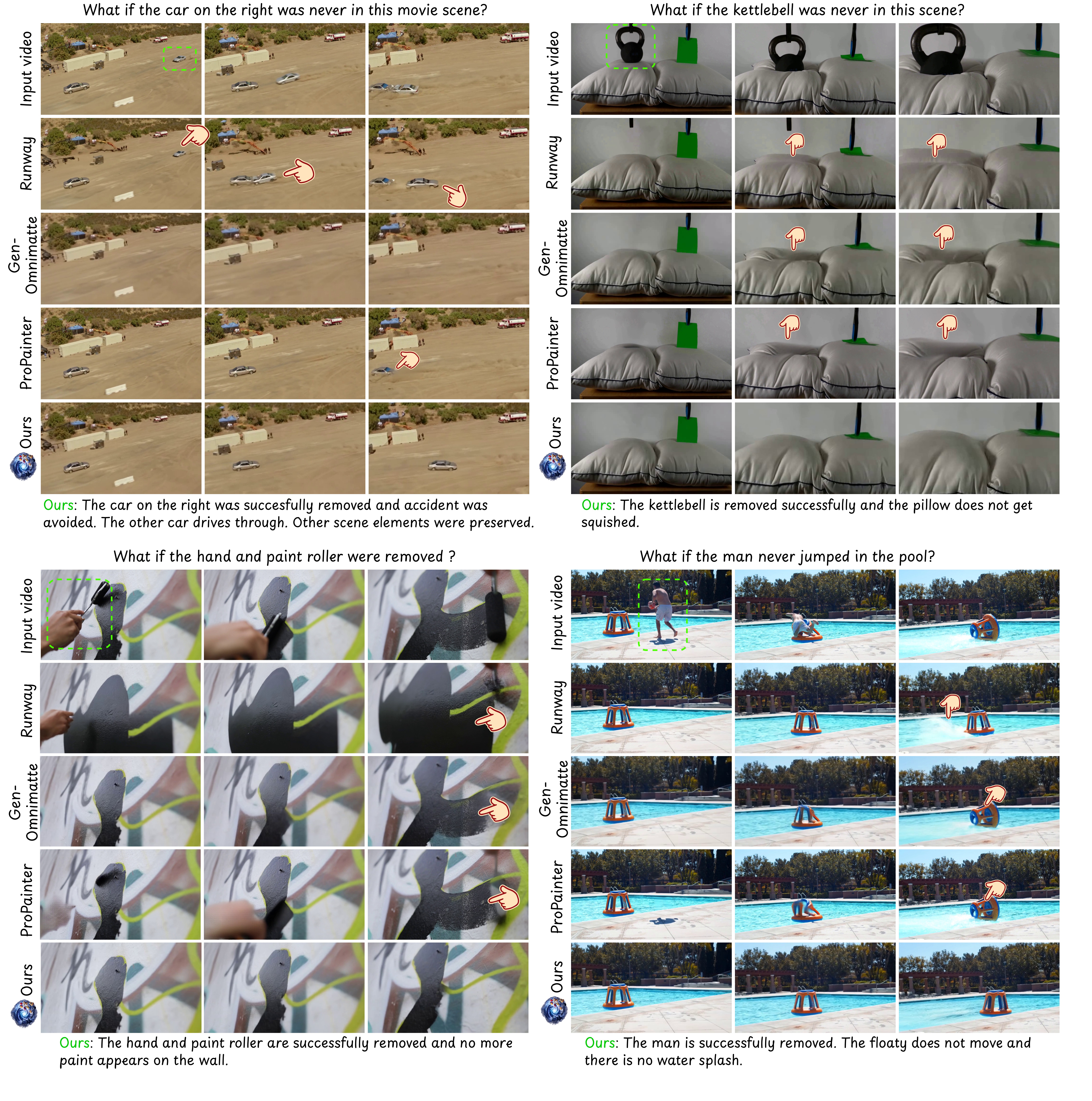}
    \caption{\textbf{Qualitative comparisons on real-world videos.}
    VOID maintains object structure and produces plausible motion over time, while the baselines exhibit deformation (the kettlebell on the pillow, and the floaty deforming), incomplete removal (two cars crashing), or implausible outputs (paint appearing after the roller is removed).
    }
    \label{fig:result-sample}
\end{figure*}

\subsection{Experimental Details}
\label{sec:eval_protocol}

For each real-world video, a user specifies a primary object via sparse clicks, which are converted into the binary object mask $\mo$ using Segment Anything 2~\cite{ravi2024sam2}. We convert the binary mask into a quadmask with the VLM-based pipeline in \cref{sec:vlm_inference}. All results in the main paper use Gemini 3 Pro as the VLM in this pipeline; we also report scores with GPT-5.2 and Qwen-3.5 VL in the appendix.

For fair comparison, each baseline is evaluated using its preferred conditioning format: binary masks for ProPainter, DiffuEraser, ROSE, and MiniMax-Remover; trimasks for Generative Omnimatte; and natural-language editing prompts for Runway (Aleph), a commercial video editing system.

Since Runway is a text-guided editor rather than a mask-conditioned inpainting model, we explicitly describe both \textit{(i)} the object to remove and \textit{(ii)} the expected scene evolution after removal (e.g., ``remove the person and ensure the held object falls naturally''). This makes the counterfactual requirement explicit while allowing each model to operate under its intended interface. We did not compare with Object-Wiper~\cite{objectwiper2026} and DynaEdit~\cite{kulikov2026versatile} as code is unavailable, nor OmnimatteZero~\cite{samuel2025omnimattezero} due to an acknowledged issue with their released code at the time of writing.

\subsection{Real-World Counterfactual Comparisons}
Since there are no ground truth counterfactuals for real-world videos, we evaluate with a human preference study, three VLM judges on fine-grained criteria, and several qualitative comparisons.

\paragraph{\textbf{Human Preference Study}.}
\label{sec:user_study}

\setlength{\intextsep}{0pt}     
\setlength{\columnsep}{4pt}
\setlength{\belowcaptionskip}{0pt}
\setlength{\abovecaptionskip}{2pt}

\begin{figure*}[ht!]
    \centering
    \includegraphics[width=\linewidth]{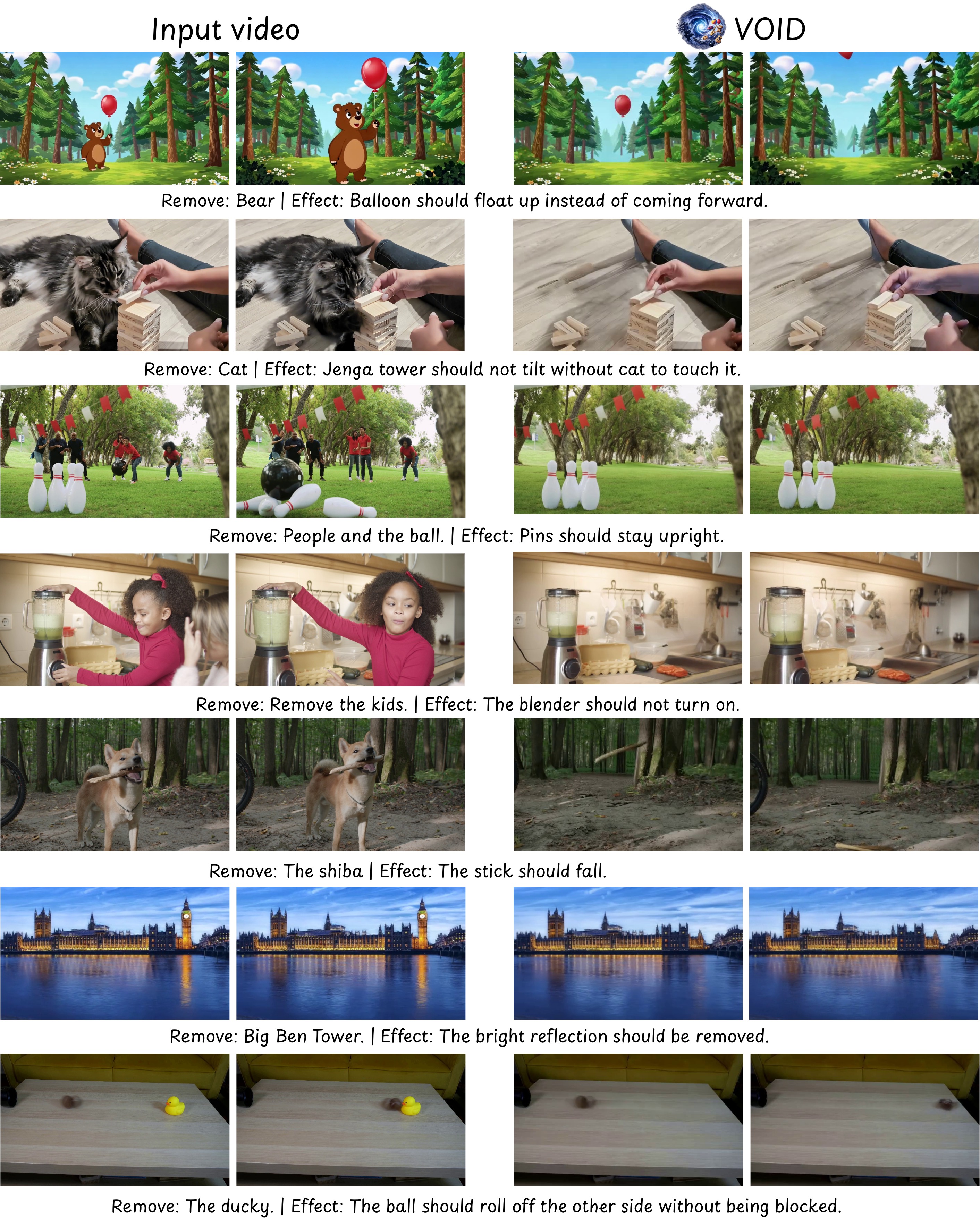}
    \caption{\textbf{Generalization on various object interactions.}
    VOID removes the target object and resolves downstream physical consequences (e.g., released objects fall; prevented collisions do not occur; shadows/reflections disappear).}
    \label{fig:more-results}
\end{figure*}
\begin{table}
  \centering

\begin{tabular}{lc}
\hline
\textbf{Model} & \textbf{Win \%} \\
\hline
\methodname{} (ours)  & \textbf{64.8} \\
Runway           & 18.4 \\
Gen-Omni.    & 11.2 \\
DiffuEraser      & 4.0 \\
ROSE             & 1.6 \\
MiniMax-Rem.  & 0.0 \\
ProPainter       & 0.0 \\
\hline
\end{tabular}
  \caption{\textbf{Human preferences on real-world edits.} 25 participants each evaluated 5 scenarios.}
  \label{tab:user_study_results}
\end{table}

We conduct a user study with 25 participants to measure perceptual realism and physical plausibility of counterfactual edits. For each participant, we randomly sample 5 out of the 75 real-world scenarios, resulting in 125 total comparisons. For each video, participants saw the original input and outputs of all seven models in randomized order. They are asked to select the video that best reflects how the scene should \emph{realistically} appear after the specified object is removed, considering visual quality, temporal consistency, blending, realism of scene evolution, and absence of artifacts. An example of the user interface for the user study is provided in the appendix.

Table~\ref{tab:user_study_results} summarizes the results. VOID is selected \textbf{64.8\%} of the time, substantially outperforming all baselines, including the closed-source Runway model which required additional text guidance on what should happen. Models optimized for traditional inpainting (e.g., ProPainter) receive few or no selections, showing that they are not automatically capable of interaction-aware synthesis.

\paragraph{\textbf{VLM-as-a-judge evaluation.}}
\label{sec:vlm_judge}

To complement human evaluation with more fine-grained criteria, we employ three VLMs (Gemini 3 Pro, GPT-5.2, and Qwen 3.5-32B) as automated judges~\cite{xiong2025llava,chen2024mllm}. Each judge scores outputs across six criteria (0--5 per category; total 30): ``Interaction \& Physics'', ``Object Removal'', ``Background \& Artifacts'', ``Temporal Consistency'', ``Preservation'', and ``Sharpness''. Table~\ref{tab:model_comparison} reports the full results.
Across all three judges, VOID achieves the highest total score. The overall ranking is broadly consistent across judges and aligns with the human preference study: VOID is ranked first, Runway second, and Generative Omnimatte third in most cases. The strongest and most consistent gains appear in ``Interaction \& Physics'', which directly evaluates whether the scene updates causally after removal. VOID correctly simulates the intuitive physics while achieving visual quality at least on par with the Runway general video editing model.

\begin{table*}[t]
\centering
\caption{\textbf{VLM-as-a-judge evaluation on real-world videos.}
Each criterion is scored in $[0,5]$. Best per judge and column is \textcolor{bestgreen}{green}; second best is \textcolor{secondorange}{orange}.}
\label{tab:model_comparison}


\begin{tabular}{@{}cl
                S[table-format=1.2]
                S[table-format=1.2]
                S[table-format=1.2]
                S[table-format=1.2]
                S[table-format=1.2]
                S[table-format=1.2]
                S[table-format=2.2]@{}}
\toprule
\textbf{Judge} & \textbf{Model} &
{Int.Phys$\uparrow$} &
{Obj.Rem$\uparrow$} &
{Bg.Art$\uparrow$} &
{Temp$\uparrow$} &
{Pres$\uparrow$} &
{Sharp$\uparrow$} &
\textbf{Total$\uparrow$} \\
\midrule

\multirow{7}{*}{\rotatebox{90}{\textbf{Gemini-3 Pro}}}
& ProPainter      & 0.82 & 3.81 & 2.64 & 3.64 & 4.81 & 3.35 & 19.05 \\
& DiffuEraser     & 1.48 & 4.43 & 3.53 & 4.12 & 4.84 & 4.13 & 22.53 \\
& MiniMax-Remover & 1.94 & 4.47 & 3.70 & 4.30 & 4.80 & \cellcolor{lightorange}4.24 & 23.46 \\
& ROSE            & 2.25 & \cellcolor{lightorange}4.77 & 3.86 & 4.26 & \cellcolor{lightgreen}4.92 & 4.17 & 24.22 \\
& Gen-Omnimatte   & 2.30 & 4.75 & 3.81 & 4.32 & \cellcolor{lightgreen}4.92 & 4.23 & 24.34 \\
& Runway          & \cellcolor{lightorange}2.61 & 4.62 & \cellcolor{lightgreen}4.16 & \cellcolor{lightgreen}4.49 & 4.82 & \cellcolor{lightgreen}4.35 & \cellcolor{secondorange}25.05 \\
& Ours            & \cellcolor{lightgreen}3.66 & \cellcolor{lightgreen}4.82 & \cellcolor{lightorange}4.10 & \cellcolor{lightorange}4.44 & \cellcolor{lightorange}4.88 & 4.22 & \cellcolor{bestgreen}26.13 \\
\midrule

\multirow{7}{*}{\rotatebox{90}{\textbf{GPT-5.2}}}
& ProPainter      & 0.76 & 2.91 & 2.11 & 2.44 & 3.13 & 3.20 & 14.55 \\
& MiniMax-Remover & 0.81 & 3.49 & 2.51 & 2.71 & 3.17 & 3.67 & 16.31 \\
& DiffuEraser     & 0.97 & 3.53 & 2.67 & 2.79 & 3.32 & 3.64 & 17.00 \\
& ROSE            & 1.39 & \cellcolor{lightorange}4.13 & 2.95 & 2.93 & 3.51 & 3.67 & 18.59 \\
& Gen-Omnimatte   & 1.33 & 4.05 & 3.25 & 3.31 & 3.56 & 3.73 & 19.23 \\
& Runway          & \cellcolor{lightorange}1.85 & 3.59 & \cellcolor{lightgreen}3.60 & \cellcolor{lightorange}3.48 & \cellcolor{lightorange}3.75 & \cellcolor{lightgreen}3.89 & \cellcolor{secondorange}20.21 \\
& Ours            & \cellcolor{lightgreen}3.19 & \cellcolor{lightgreen}4.35 & \cellcolor{lightorange}3.48 & \cellcolor{lightgreen}3.88 & \cellcolor{lightgreen}4.41 & \cellcolor{lightorange}3.81 & \cellcolor{bestgreen}23.16 \\
\midrule

\multirow{7}{*}{\rotatebox{90}{\textbf{Qwen3.5-32B}}}
& ProPainter      & 1.93 & 4.43 & 2.96 & 3.89 & \cellcolor{lightgreen}5.00 & 4.00 & 22.21 \\
& MiniMax-Remover & 1.68 & 4.60 & 3.12 & 4.03 & \cellcolor{lightgreen}5.00 & 4.03 & 22.45 \\
& Runway          & 2.03 & 4.25 & \cellcolor{lightorange}3.36 & 4.07 & 4.95 & \cellcolor{lightorange}4.05 & 22.79 \\
& ROSE            & 1.75 & 4.45 & 3.41 & \cellcolor{lightorange}4.17 & 4.93 & 4.04 & 22.96 \\
& DiffuEraser     & \cellcolor{lightorange}2.19 & \cellcolor{lightorange}4.75 & 3.04 & 4.00 & \cellcolor{lightgreen}5.00 & 4.00 & 23.04 \\
& Gen-Omnimatte   & \cellcolor{lightorange}2.19 & \cellcolor{lightgreen}4.89 & 3.16 & 4.08 & \cellcolor{lightgreen}5.00 & \cellcolor{lightorange}4.05 & \cellcolor{secondorange}23.44 \\
& Ours            & \cellcolor{lightgreen}2.64 & \cellcolor{lightorange}4.75 & \cellcolor{lightgreen}3.55 & \cellcolor{lightgreen}4.24 & \cellcolor{lightgreen}5.00 & \cellcolor{lightgreen}4.12 & \cellcolor{bestgreen}24.49 \\
\bottomrule
\end{tabular}
\end{table*}

\paragraph{\textbf{Qualitative comparisons.}} \cref{fig:result-sample} presents qualitative comparisons on 4 real-world videos. We present 3 representative baselines: Runway for text-based video editing, Gen-Omnimatte for effect removal inpainting, and Propainter for traditional video inpainting without effect removal. All baselines present various types of failure cases, such as not removing anything or removing more than intended (two-car crashing example) or creating physically implausible scenes (pillow squished without kettlebell, floaty falling without a collision, and paint still appearing after the paint roller was removed). \methodname{} exhibits high generalization capabilities across all examples, performing accurate object and effect removal while ensuring the scene remains physically plausible and artifact-free.


\paragraph{\textbf{Generalizations to unseen effects.}} \Cref{fig:more-results} shows generations by \methodname{} on samples from our real-world dataset involving effects unseen in training. To our delight, \methodname{} is frequently able to extrapolate to these new types of physical interactions. It disentangles complex motions, such as a Jenga tower being simultaneously pushed by a hand and a cat, and a bowling ball hitting multiple bowling pins. It infers physics effects not present in the training dataset, such as a balloon floating up after the holder is removed, and a blender not turning on when the person pressing it is removed. Finally, it remains robust to removing the reflection of the Big Ben tower, letting the stick fall when the dog chewing it is removed, and corrects the ball rolling trajectory when the ducky obstacle is removed. The diverse set of effects are strong indicators that \methodname{} learns to leverage the intuitive physics reasoning of the VLM and video diffusion base model in a general manner, letting it excel on tasks far from the synthetic data we use to train it. We provide more video examples in the appendix.

\subsection{Synthetic Dataset Comparisons}
\label{sec:synthetic_eval}

To compute metrics requiring ground-truth counterfactual targets, we take a synthetic benchmark of 10 videos focusing on object/shadow/reflection removal used by prior work~\cite{lee2024generative} and add another 30 dynamic counterfactual cases from Kubric and HUMOTO that captures a wider range of object interactions. These include altered collision outcomes and released objects entering free fall. These videos were held-out from our training dataset.

\begin{table*}[!t]
\centering
\caption{\textbf{Synthetic benchmark evaluation}
on 10 classic shadow/reflection removal cases and 30 dynamic interaction cases (Kubric + HUMOTO) capturing a wide range of effects. 
All metrics measure fidelity to the ground-truth counterfactual targets.}
\label{tab:synthetic_benchmark}


\begin{tabular}{@{}l
S[table-format=2.2]
S[table-format=1.4]
S[table-format=1.4]
S[table-format=1.4]
S[table-format=3.2]
S[table-format=2.2]@{}}
\toprule
\textbf{Model} &
\textbf{PSNR$\uparrow$} &
\textbf{LPIPS$\downarrow$} &
\textbf{DreamSim$\downarrow$} &
\textbf{DINOv2$\uparrow$} &
\textbf{FVD$\downarrow$} &
\textbf{VLM-Judge$\uparrow$} \\
\midrule

MiniMax-Remover &
29.96 &
\cellcolor{secondorange}0.1091 &
\cellcolor{secondorange}0.0879 &
\cellcolor{secondorange}0.91 &
448.43 &
\cellcolor{secondorange}22.83 \\

ProPainter &
\cellcolor{secondorange}30.48 &
\cellcolor{bestgreen}0.0981 &
0.0987 &
0.8881 &
471.13 &
21.38 \\

DiffuEraser &
30.11 &
0.1213 &
0.0992 &
0.8920 &
496.61 &
21.30 \\

ROSE &
29.21 &
0.1296 &
0.1104 &
0.8905 &
480.18 &
21.62 \\

Gen-Omnimatte &
29.44 &
0.1186 &
0.1219 &
0.8688 &
\cellcolor{secondorange}437.88 &
20.40 \\

Runway &
26.68 &
\cellcolor{secondorange}0.11 &
0.1464 &
0.8522 &
442.76 &
21.67 \\

\midrule

\textbf{Ours} &
\cellcolor{bestgreen}31.49 &
0.1156 &
\cellcolor{bestgreen}0.0658 &
\cellcolor{bestgreen}0.9222 &
\cellcolor{bestgreen}260.31 &
\cellcolor{bestgreen}25.10 \\

\bottomrule
\end{tabular}
\end{table*}

We follow previous work in reporting pixel-based metric PSNR and perceptual metric LPIPS~\cite{lpips}. However, with the new dataset introducing more diverse sets of effects, we also add in the more recent frame-wise perceptual metrics DreamSim~\cite{fu2023dreamsim} and DINOv2~\cite{oquab2023dinov2}, as well as video metric FVD~\cite{unterthiner2019fvd}. These can better capture intricate and semantically high-level effects. We also include a VLM-Judge evaluation by Gemini 3 Pro, which is the closest aligned with the human evaluation on our real-world dataset. We modify our VLM-judge protocol on this dataset by showing the judge the ground-truth counterfactual video in addition to the model output. The same six-category scoring protocol (0--5 per criterion; total 30) is applied as in the real-world setting, but judges now assess fidelity relative to the true counterfactual outcome.

VOID achieves the strongest performance across all metrics except LPIPS. Note that this frame-wise LPIPS metric is sensitive to local translations, and therefore can penalize counterfactual effects being generated in slightly incorrect regions. For example, if we remove a person holding a stick, a model accurately portraying a stick falling but at the wrong speed may be penalized more than a model that removes the stick altogether.
The largest margin between \methodname{} and baselines appears in the FVD and the VLM-judge metrics, which are the two comprehensive video-level metrics we report. This is strongly supportive of our claim that \methodname{} excels at producing physically plausible and semantically coherent videos.

\subsection{Ablation}
\label{sec:ablation}

\Cref{tab:ablation} presents ablations on our training data and quadmask strategy. 
To best capture model robustness and generalization, all variants are evaluated on the same 75 real-world test cases using Gemini 3 Pro as a VLM judge. See the appendix for a further ablation on VOID's second pass.

\paragraph{\textbf{Data composition.}}
To analyze the effect of our datasets, we train ablations with three alternative datasets: Kubric-Only is trained on a 1200 sample subset of Kubric; HUMOTO-Only is trained on a 1200 sample subset of HUMOTO; and Both Datasets is trained on another 1200 samples split equally between Kubric and HUMOTO. In \cref{tab:ablation} we see that Kubric-Only and HUMOTO-Only both underperform Both-Datasets, meaning the diversity we get by mixing Kubric and HUMOTO data is beneficial even when the dataset size is held constant.

\paragraph{\textbf{Masking strategy.}}
We train another ablation that uses less detailed trimasks, similar to Generative Omnimatte~\cite{lee2024generative}, so that we can drop the VLM-guided mask generation pipeline. These masks are simply black wherever the object to be removed is and light gray everywhere else, meaning that there are no constraints on what parts of the video the diffusion model can change. \Cref{tab:ablation} shows that this ablation degrades performance across all categories, confirming the importance of our detailed masks and our mask generation pipeline.

\begin{table*}[t]
\centering
\caption{\textbf{Ablation study} evaluated by VLM judge on 75 real-world test cases.}
\label{tab:ablation}

\renewcommand{\arraystretch}{1.1}

\begin{tabular}{@{}lccccccc@{}}
\toprule
\textbf{Model (Dataset Size)} &
{Int.Phys$\uparrow$} &
{Obj.Rem$\uparrow$} &
{Bg.Art$\uparrow$} &
{Temp$\uparrow$} &
{Pres$\uparrow$} &
{Sharp$\uparrow$} &
\textbf{Total$\uparrow$} \\
\midrule

Kubric-Only (1200)&
2.63 &
4.06 &
2.33 &
3.46 &
4.34 &
3.54 &
20.36 \\

HUMOTO-Only (1200) &
2.50 &
4.22 &
2.36 &
3.33 &
4.30 &
3.41 &
20.12 \\

Both Datasets (1200)&
3.04 &
4.30 &
2.31 &
3.87 &
4.45 &
3.96 &
21.93 \\

Gen-Omni. Mask (Full)&
3.30 &
4.73 &
3.04 &
4.04 &
4.22 &
4.06 &
23.39 \\

\midrule

VOID (Full) &
3.66 &
4.82 &
4.10 &
4.44 &
4.88 &
4.22 &
26.12 \\

\bottomrule
\end{tabular}
\end{table*}

\section{Conclusion}
We present \methodname{}, an object removal framework that generates the counterfactual video corresponding to when an object is removed. \methodname{} is built upon two new paired datasets of counterfactual object removal videos derived from the Kubric engine and HUMOTO dataset. We also present a VLM-guided quadmask generation pipeline to guide \methodname{} into generating physics-informed counterfactual videos. Through extensive evaluations against inpainting and text-guided video model baselines on synthetic and real-world data, we show that \methodname{} excels at modeling complex dynamics which can follow on from object removal. It also generalizes to a broad range of scenarios far from our training data. \methodname{} is a strong starting point for future research to continue transferring strong world modeling capabilities to the video editing domain.

\paragraph{\textbf{Limitations and future work.}}
Despite the various generalization capabilities \methodname{} exhibits, there are still certain domain gaps we observe, such as when test videos have the cameras at an unusual angle or too close to the object. Future work could obtain better training datasets beyond rendering engines. The generated video lengths are still in the range of a few seconds, and resolutions could be further improved.
\label{sec:conclusion}


\setcounter{section}{0}
\renewcommand{\thesection}{(\roman{section})}
\FloatBarrier
\bibliographystyle{splncs04}
\bibliography{main}
\clearpage
\onecolumn

\setcounter{section}{0}
\renewcommand{\thesection}{(\roman{section})}

\begin{center}
\LARGE \textbf{VOID: Supplementary Material}
\end{center}

Below we provide additional experimental details and analyses supporting the main paper. 
Specifically, we present (i) an analysis of mask generation using different VLM reasoners, (ii) results for the second-pass refinement stage, (iii) the user interfaces used for mask generation and (iv) human evaluation, (v) examples illustrating limitations of standard video similarity metrics, and (vi) the full prompts used for the VLM-based evaluation protocol.

\section{Mask Generation with Different VLM Reasoners}
\label{supp:mask_generation}
Our pipeline uses a vision-language model (VLM) to generate interaction-aware masks from sparse user input. To study how the choice of VLM affects mask quality, we evaluate three models: Qwen3-32B, GPT~5.2, and Gemini~3-Pro. These models receive identical user clicks and produce masks that guide the inpainting process.

Table~\ref{tab:supp_mask_eval} reports Gemini~3 VLM-judge scores measuring the quality of the resulting inpainted videos across six dimensions. The judge evaluates interaction physics, object removal, background artifacts, temporal consistency, preservation of the scene, and sharpness.

\begin{table*}[!bp]
\centering
\caption{Gemini~3 VLM-judge evaluation on 75 real-world videos when \methodname{} uses different VLMs for mask generation during inference.}
\label{tab:supp_mask_eval}

\scriptsize
\setlength{\tabcolsep}{3.5pt}

\begin{tabular}{lccccccc}
\toprule
\textbf{Mask Reasoner} &
\textbf{Int.Phys} &
\textbf{Obj.Rem} &
\textbf{Bg.Art} &
\textbf{Temp} &
\textbf{Pres} &
\textbf{Sharp} &
\textbf{Total} \\
\midrule

Qwen3-32B
& \cellcolor{bestgreen}3.75
& \cellcolor{lightorange}4.23
& 3.54
& \cellcolor{lightorange}4.11
& 4.86
& 3.42
& 23.91 \\

GPT~5.2
& 3.49
& 4.17
& \cellcolor{lightorange}3.60
& 4.03
& \cellcolor{bestgreen}4.94
& \cellcolor{lightorange}4.11
& \cellcolor{lightorange}24.34 \\

Gemini 3-Pro
& \cellcolor{lightorange}3.66
& \cellcolor{bestgreen}4.82
& \cellcolor{bestgreen}4.10
& \cellcolor{bestgreen}4.44
& \cellcolor{lightorange}4.88
& \cellcolor{bestgreen}4.22
& \cellcolor{bestgreen}26.12 \\

\bottomrule
\end{tabular}
\end{table*}

Gemini~3-Pro consistently produces the most reliable masks, particularly improving interaction physics and background reconstruction.

\section{Second-Pass Refinement Analysis}
\label{supp:pass2}

VOID optionally performs a second refinement pass when the VLM determines that the object-interaction removal requires substantial reconfiguration of scene elements to produce a physically plausible outcome. Among the 75 real-world test videos, the VLM flagged 10 cases as requiring refinement. Table~\ref{tab:vlm-eval-categories} compares Pass~1 and Pass~2 results on these samples using the VLM judge.

\begin{table*}[!bp]
\centering
\caption{Per-category average scores (out of 5) across the 10 samples that are selected by the VLM for pass 2 refinement.}
\label{tab:vlm-eval-categories}

\scriptsize
\setlength{\tabcolsep}{4pt}

\begin{tabular}{lccccccc}
\toprule
\textbf{Pass} &
\textbf{Int.Phys} &
\textbf{Obj.Rem} &
\textbf{Bg.Art} &
\textbf{Temp} &
\textbf{Pres} &
\textbf{Sharp} &
\textbf{Total} \\
\midrule

Pass 1
& 2.90
& 4.20
& 3.70
& 3.80
& \cellcolor{bestgreen}4.90
& 4.00
& 23.5 \\

Pass 2
& \cellcolor{bestgreen}3.90
& \cellcolor{bestgreen}4.90
& \cellcolor{bestgreen}4.00
& \cellcolor{bestgreen}4.20
& 4.80
& \cellcolor{bestgreen}4.20
& \cellcolor{bestgreen}26.0 \\

\bottomrule
\end{tabular}
\end{table*}

The refinement step improves interaction reasoning and object removal quality, leading to higher overall scores.

\section{User Interface for Mask Generation}
\label{supp:mask_ui}

\begin{figure*}[!tbp]
\centering
\includegraphics[width=\linewidth]{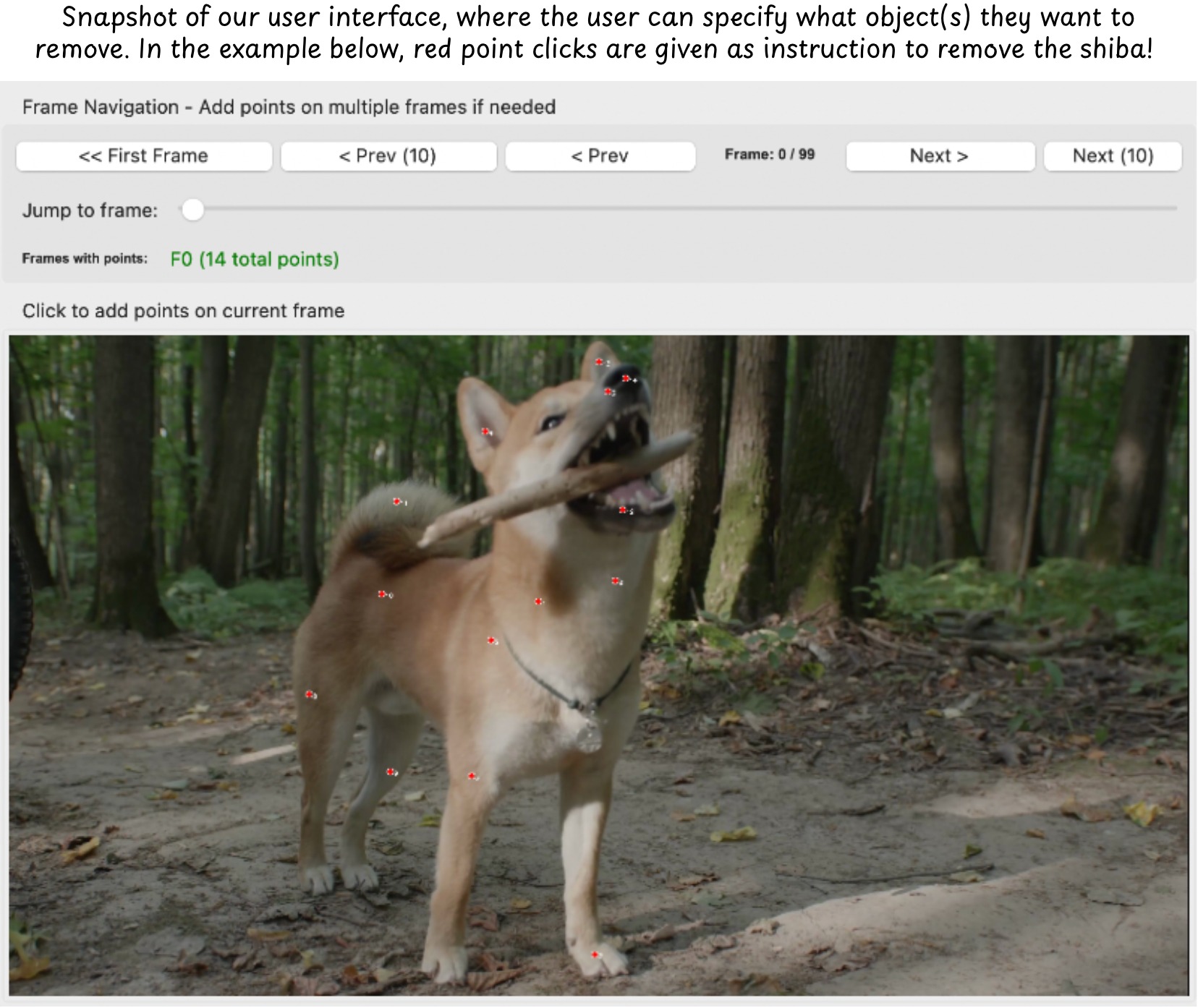}
\caption{User interface for mask generation. The user selects sparse points on the target object, and the VLM generates the removal mask.}
\label{fig:mask_ui}
\end{figure*}

Figure~\ref{fig:mask_ui} shows the graphical interface used to collect sparse user inputs. Users select a small number of points on the object to be removed, and the VLM generates an interaction-aware mask conditioned on the scene context.

\section{User Study Interface}
\label{supp:user_study}

\begin{figure*}
\centering
\includegraphics[width=\linewidth]{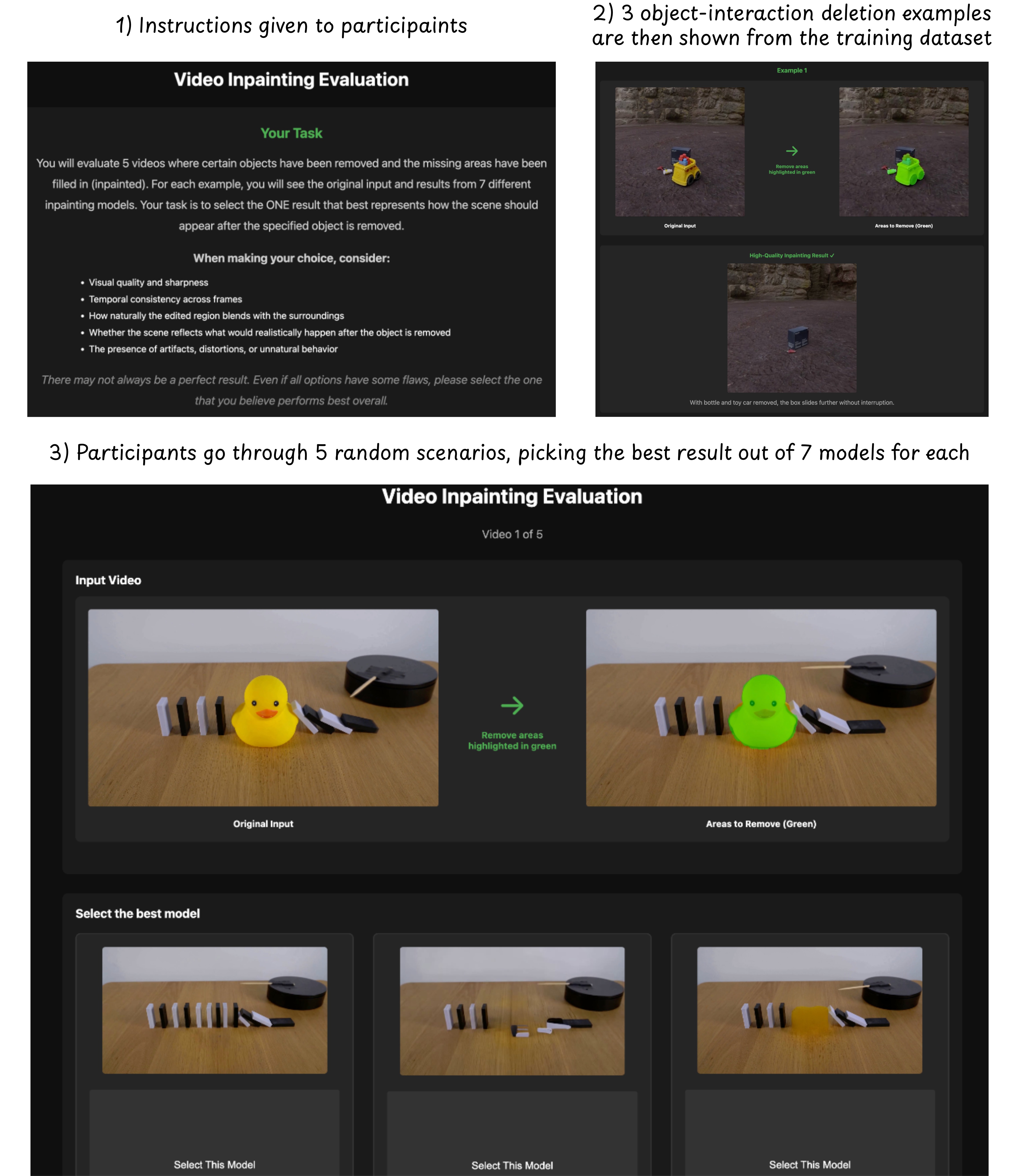}
\caption{Interface used in the human evaluation study.}
\label{fig:userstudyui}
\end{figure*}

Figure~\ref{fig:userstudyui} shows the interface used in our human evaluation study. Participants first read a brief set of instructions and are shown three example object-interaction removal cases from the training dataset. They are then presented with five randomly sampled scenarios for evaluation. For each scenario, participants see the original input video as well as the same video with the objects to be removed highlighted in green. Participants can then view the outputs of seven different models (VOID, Runway, Generative Omnimatte, DiffuEraser, ProPainter, MiniMax-Remover, ROSE) and select the result that produces the best inpainting outcome.

\section{Limitations of Standard Video Metrics}
\label{supp:metrics}

\begin{figure*}
\centering
\includegraphics[width=\linewidth]{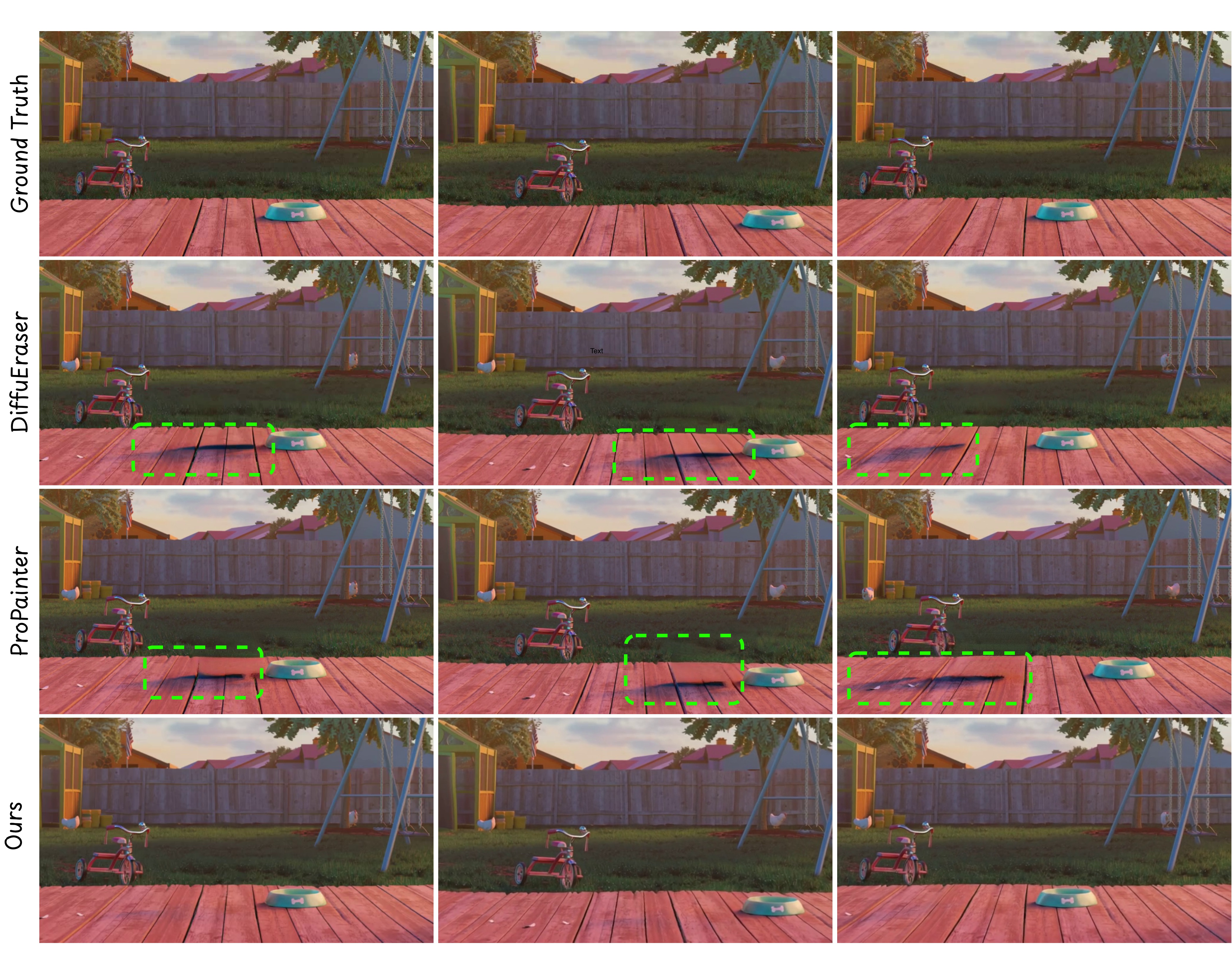}
\caption{Examples where standard video similarity metrics favor visually implausible results.}
\label{fig:metric_failures}
\end{figure*}

Standard perceptual similarity metrics such as LPIPS, DreamSim, and feature-based similarity measures (e.g., DINOv2) are widely used for evaluating visual fidelity and perceptual similarity. While these metrics are valuable indicators of image or video similarity, they may fail to capture certain task-specific artifacts relevant to video inpainting, particularly in dynamic settings involving object interactions and causal effects. In some cases, methods that produce visually implausible or blurry results can achieve better scores than models with objectively more realistic inpainting outcomes. Figure~\ref{fig:metric_failures} shows examples where DiffuEraser~\cite{li2025diffueraser} and ProPainter~\cite{zhou2023propainter} obtain better LPIPS, DreamSim and DINOv2 scores (Table~\ref{tab:dog_metrics}) despite leaving a clear shadow artifact of the removed object in the scene. This illustrates that similarity-based metrics primarily measure appearance-level correspondence and may overlook physically implausible artifacts or incorrect scene dynamics that are particularly important for interaction-aware video editing tasks.

\begin{table*}[!tbp]
\centering
\begin{tabular}{lcccc}
\toprule
Model & LPIPS$\downarrow$ & PSNR$\uparrow$ & DreamSim$\downarrow$ & DINOv2$\uparrow$ \\
\midrule
ProPainter   & 0.0879 & 27.98 & 0.0459 & 0.9824 \\
DiffuEraser  & \textbf{0.0823} & 28.14 & \textbf{0.0384} & \textbf{0.9836} \\
Ours         & 0.1081 & \textbf{28.96} & 0.0486 & 0.9786 \\
\bottomrule
\end{tabular}
\caption{Metrics calculated for the example shown in Figure~\ref{fig:metric_failures}.}
\label{tab:dog_metrics}
\end{table*}
\newpage
\section{VLM Judge Prompt}
\label{supp:judge_prompt}

Below, we provide the full prompts used to instruct the VLM judge in evaluating inpainting quality of the videos by giving a score of \(0 - 5\) in 6 categories for a total score of $30$.

\paragraph{Stage 1 Prompt.}
The VLM receives the \emph{original} video and returns a structured scene understanding used as context in Stage~2.

\begin{codeblock}
INPUT TO VLM
============
[Video] original, unedited input video
[Text]  (shown below)

Watch this ORIGINAL video and analyze the removal instruction:
"<removal_instruction>"

VIDEO INPAINTING WITH INTERACTION AWARENESS means understanding the CAUSAL PHYSICS: if a person holding a mug is removed, the mug should FALL; if a ball knocked things over, removing the ball means those things should NOT fall; if someone casts a shadow, removing them should remove the shadow.

Analyze the following:
1. What object/subject should be removed and what is it physically interacting with (holding, pushing, casting shadows, etc.)?
2. What objects are supported, moved, or otherwise affected by the target, i.e. what are the physical consequences of its removal?
3. What should the background look like after perfect removal?
4. What visual effects (shadows, reflections) must also disappear?

OUTPUT (JSON)
=============
{
  "target_object": "...",
  "object_interactions": ["...", "..."],
  "physical_consequences": ["...", "..."],
  "expected_background": "...",
  "visual_effects_to_remove": ["shadow", "reflection", "..."],
  "should_not_change": "...",
  "interaction_difficulty": "easy / medium / hard",
  "interaction_difficulty_reasoning": "..."
}
\end{codeblock}

\vspace{6pt}

\paragraph{Stage 2 Prompt.}
The VLM receives an \emph{inpainting result} video together with the structured context produced in Stage~1 and returns per-dimension scores (0–5) summing to a maximum of 30.

\begin{codeblock}
INPUT TO VLM
============
[Video] inpainting result video (model output to be scored)
[Text]  (shown below; <...> fields filled from Stage 1 output)

CONTEXT FROM SCENE UNDERSTANDING
================================
Removal instruction: "<removal_instruction>"
Target object: <target_object>
Interactions: <object_interactions>

Physical consequences that MUST be present:
  * <consequence_1>
  * <consequence_2>

Visual effects to remove: <visual_effects_to_remove>
Background to fill: <expected_background>
Must NOT change: <should_not_change>

YOUR TASK
=========
Watch the inpainting result video and score it on the six dimensions below (0–5 each, max 30 total). Pay close attention to object MOVEMENTS and TRAJECTORIES. Only report motion you clearly observe frame-by-frame; do not hallucinate motion in stationary objects.

SCORING DIMENSIONS
==================
1. Interaction Physics [PRIMARY]
   Does the result correctly handle the physical consequences of removal
   (object trajectories, gravity, momentum, shadow/reflection removal)?
   5 - All consequences handled correctly
   4 - Mostly correct, minor physics imperfections
   3 - Partial: some consequences correct, others not
   2 - Major physics violations
   1 - Almost no interactions handled
   0 - Complete failure; interactions entirely ignored

2. Object Removal Quality
   5 - Complete removal, no traces
   3 - Removed but with noticeable artifacts/remnants
   0 - Not removed at all

3. Background & Artifact Quality
   5 - Perfectly natural, no artifacts
   3 - Acceptable with noticeable artifacts
   0 - Completely unrealistic

4. Temporal Consistency
   5 - Perfect frame-to-frame consistency
   3 - Noticeable but acceptable flickering
   0 - Completely inconsistent

5. Preservation of Scene
   5 - Non-target areas perfectly preserved
   3 - Noticeable unwanted modifications
   0 - Scene completely altered

6. Sharpness / Blur
   5 - Perfectly sharp
   3 - Slight blur but acceptable
   0 - Completely blurred / unusable

OUTPUT (JSON)
=============
{
  "scores": {
    "interaction_physics":  {"score": 0-5, "reasoning": "..."},
    "object_removal":       {"score": 0-5, "reasoning": "..."},
    "background_artifacts": {"score": 0-5, "reasoning": "..."},
    "temporal_consistency": {"score": 0-5, "reasoning": "..."},
    "preservation":         {"score": 0-5, "reasoning": "..."},
    "sharpness":            {"score": 0-5, "reasoning": "..."}
  },
  "total_score": "0-30",
  "overall_assessment": "...",
  "strengths": ["...", "..."],
  "weaknesses": ["...", "..."]
}
\end{codeblock}

\end{document}